\documentclass[sigconf,nonacm]{acmart}

\AtBeginDocument{%
  }

\setcopyright{none}
\renewcommand\footnotetextcopyrightpermission[1]{} % Removes copyright footnote
\usepackage{subfigure}
\usepackage{xcolor}
\usepackage{colortbl}
\usepackage{array}
\usepackage{caption}
\usepackage{{booktabs}}

\definecolor{blackcell}{HTML}{000004}
\definecolor{orangecell}{HTML}{f98e09}
\definecolor{redcell}{HTML}{bc3754}
\definecolor{greencell}{HTML}{249d53}
\definecolor{yellowcell}{HTML}{fcffa4}
\definecolor{lighttext}{HTML}{f1f1f1}
\definecolor{lightorangecell}{HTML}{fbb61a}
\definecolor{darkorangecell}{HTML}{ed6925}

\begin{document}

%\title{Reasoning Promotes Robustness in ToM Tasks}
\title{Reasoning Promotes Robustness in Theory of Mind Tasks}

\author{Ian B. de Haan}
\orcid{0009-0009-8074-8265}
\email{i.b.de.haan.2@umail.leidenuniv.nl}
\affiliation{%
  \institution{LIACS, Leiden University}
\city{Leiden}
  \country{The Netherlands}
}

\author{Peter van der Putten}
\orcid{0000-0002-6507-6896}
\email{p.w.h.van.der.putten@liacs.leideniniv.nl}
\affiliation{%
  \institution{LIACS, Leiden University}
\city{Leiden}
  \country{The Netherlands}}

\author{Max van Duijn}
\orcid{0000-0003-0798-9598}
\email{m.j.van.duijn@liacs.leidenuniv.nl}
\affiliation{%
\institution{LIACS, Leiden University}
\city{Leiden}
  \country{The Netherlands}
}

\renewcommand{\shortauthors}{de Haan al.}


\begin{abstract}
Large language models (LLMs) have recently shown strong performance on Theory of Mind (ToM) tests, prompting debate about the nature and true performance of the underlying capabilities. At the same time, reasoning-oriented LLMs trained via reinforcement learning with verifiable rewards (RLVR) have achieved notable improvements across a range of benchmarks. This paper examines the behavior of such reasoning models in ToM tasks, using novel adaptations of machine psychological experiments and results from established benchmarks. We observe that reasoning models consistently exhibit increased robustness to prompt variations and task perturbations. Our analysis indicates that the observed gains are more plausibly attributed to increased robustness in finding the correct solution, rather than to fundamentally new forms of ToM reasoning. We discuss the implications of this interpretation for evaluating social-cognitive behavior in LLMs.
\end{abstract}

%\keywords{Reasoning, Models, Robustness, LLMs, Language, AI}
\keywords{Theory of Mind, Reasoning models, Robustness}


\maketitle

\section{Introduction}
In recent years, large language models (LLMs) have become powerful and useful enough to become part of our day-to-day lives, and are increasingly being used in more autonomous and agentic settings \cite{plaat_agentic_2025}. Users tend to attribute human-like intentionality and reasoning to those models, even though the extent to which the models are capable of complex reasoning remains, to a large extent, an open research problem. \cite{zhaoSurveyLargeLanguage2025, rahwanMachineBehaviour2019}

Due to the complexity of these systems, even if all information about their architecture is known, it is still difficult to determine and predict their actual behavior. Considering this, researchers in a multitude of fields started investigating those systems and their intelligence by interacting with them and understanding their behavior, rather than by analyzing their architecture.  \cite{rahwanMachineBehaviour2019} 

In this context, previous work has debated the extent to which LLMs demonstrate theory of mind (ToM) behavior, that is, the ability to reason about mental states, beliefs, intentions, and desires. Some early claims were optimistic of such a skill emerging to some extent \cite{kosinski_evaluating_2024}. However, follow-up criticism \cite{ullmanLargeLanguageModels2023} led to the creation of benchmarks testing these abilities more comprehensively \cite{duijnTheoryMindLarge2023, kimFANToMBenchmarkStresstesting2023, chen2024tombench}.

Since then, so-called reasoning models, like OpenAI's GPT-5, Anthropics's Claude, and DeepSeek R1, have been released. The innovation behind these models is that they are trained, via reinforcement learning, to “think before answering", that is, they are trained to produce a complex chain of thought before answering the query, a form of inference-time scaling \cite{ReasoningModelsOpenAI, paliotta_thinking_2025, deepseek-ai_deepseek-r1_2025, plaat_reasoning_2025}. This new type of model might not only perform significantly better in ToM tasks, but the chain-of-thought itself might yield important insights about what those models are actually doing to answer the queries.

The goal of this paper is, therefore, to review the literature around reasoning models and connect this to ToM. Furthermore, additional experiments will be run to shed light on how state-of-the-art reasoning models perform relative to non-reasoning ones. Those experiments involve prompting models with established psychological tests for ToM from the \cite{duijnTheoryMindLarge2023} battery, plus newly designed prompts based on \cite{ullmanLargeLanguageModels2023}, to test model robustness. Lastly, benchmark results taken from \cite{kimHypothesisDrivenTheoryofMindReasoning2025} are analyzed to further substantiate our findings.

The remainder of this paper is structured as follows: Section \ref{Background} provides an overview of related work on the subject. Section~\ref{methods} introduces the materials and methods employed in the experiments, and Section~\ref{Results} presents the experimental results; Section~\ref{Discussion} interprets the obtained results, points out some limitations of the current work, and suggests new directions in which the research can be expanded; Section~\ref{conclusions} sums up the findings; Appendix~\ref{Appendix A} contains the newly introduced test. 

\section{Background}\label{Background}

In this section, a bibliographical review in the context of the research done is presented. The contents covered are \ref{CoT and Reas} -- a review of what chains of thought are and the recent reinforcement learning with verifiable rewards (RLVR) trained reasoning models, and \ref{ToM and LLMs} -- an introduction on the topic of ToM, how it would be a useful skill for LLMs to have, and the present research investigating to which extent they do.

\subsection{Chains of Thought and Reasoning Models} \label{CoT and Reas}

% see also https://arxiv.org/abs/2503.23037 for a wealth of refs on reasoning models -- probably too late now but for future reference (and do cite our paper;))

Traditional LLM models, based solely on next token prediction, face a series of challenges, among them are:
\begin{itemize}
    \item Their performance increases with the number of parameters and the size of the dataset \cite{kaplanScalingLawsNeural2020}, and we may run out of human-generated data between 2026-2032 \cite{villalobosWillWeRun2024}. Furthermore, current models are often under-trained \cite{hoffmannTrainingComputeOptimalLarge2022}.
    \item By default, they don't display human-interpretable thoughts
    \item It doesn't seem like increasing their size alone would yield in satisfactory performance in fields like math and logical reasoning tasks \cite{raeScalingLanguageModels2022}.
\end{itemize}

In this scenario, \cite{weiChainofThoughtPromptingElicits2023} demonstrated that simply feeding LLMs with a few-shot prompts, inducing them to respond in the form of <input, chain of thought, output>, not only generated step-by-step explanations of natural language but also increased performance in math and logical reasoning tasks.

\begin{figure*}[ht!]
    \centering
    \subfigure[]{\includegraphics[width=0.46\textwidth]{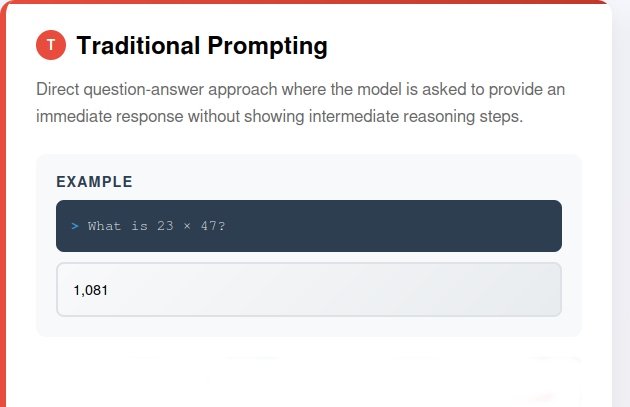}} \\
    \subfigure[]{\includegraphics[width=0.46\textwidth]{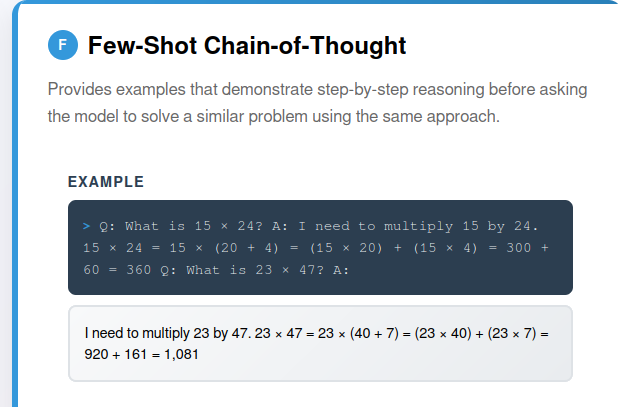}}
    \subfigure[]{\includegraphics[width=0.46\textwidth]{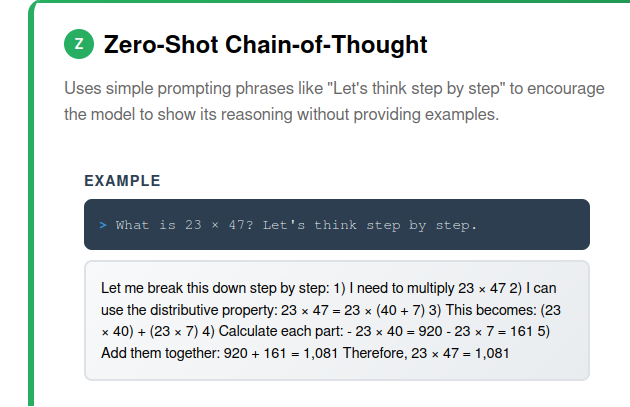}}
    \caption{Image illustrating the chain-of-thought prompting techniques (a) Traditional prompting, doesn't elicit CoT (b) Few-shot prompting, elicits CoT at the cost of a few examples (c) Zero shot prompting, elicits CoT for free}
    \label{fig:CoT prompting}
\end{figure*}

 \cite{weiChainofThoughtPromptingElicits2023} also envisioned that, on top of that, in principle, chains of thought (CoTs) could help separate a  problem into smaller tasks, allowing models to decompose problems and allocate a different amount of time to each of their sub-problems. Several variants of prompting style were proposed, notoriously simply including `let's think step by step' in the prompt, lead to LLMs producing CoT with zero-shot training \cite{kojimaLargeLanguageModels2022}. Figure \ref{fig:CoT prompting} illustrates those prompting techniques.

With CoT getting the spotlight, many different ideas on how to induce and refine the reasoning demonstrated by LLMs started to appear \cite{weiChainofThoughtPromptingElicits2023}. RLVR proved itself to be effective at inducing complex reasoning in models and, using this technique, a new wave of models, refined to produce chains of thoughts, started emerging and were denominated `reasoning models' \cite{xuLargeReasoningModels2025, plaat_reasoning_2025, plaat_agentic_2025}. The implementation details of those models are an important topic, but outside of the scope of this paper.

However, although promising results in several fields, particularly in mathematics and programming, ignited hope that RLVR-trained LLMs could achieve a more general form of intelligence, \cite{yueDoesReinforcementLearning2025} demonstrated that, in reality, the gained performance of those models comes from a more efficient sampling of answers on problems that are already solvable by the base model, not introducing fundamentally new reason capabilities. Therefore, the reasoning capacity of a reasoning model remains bounded by its base model.

Reasoning was also initially viewed as promising in the way towards tracking reward hacking in LLMs, but it has been shown that, at least in scenarios where the CoT is not necessary for such behavior, the verbalization of it is very low \cite{chenReasoningModelsDont2025}. This finding also makes one conclude, more generally, that reasoning models don't always include their true reasoning in their think tokens, although the same article also showed that, at least part of the time, they do.

\subsection{Theory of Mind and LLMs} \label{ToM and LLMs}

The term “theory of mind" first appears in the article \cite{premackDoesChimpanzeeHave1978}, where it refers to the ability of one to attribute mental states to another. The article investigates whether chimpanzees have this ability. Since such  an article, ToM started to be widely investigated and, with the attention, different definitions of it emerged. Some of those definitions, as per \cite{schlingerTheoryMindOverview2009}, are:

\begin{quote}
\itshape
Theory of mind refers to the ability to represent, conceptualize, and reason about mental states. In its fully mature stage, theory of mind is a domain-specific conceptual framework that treats certain perceptual input as an agent, an intentional action, a belief, and so forth. . . . Theory of mind arguably underlies all conscious and unconscious cognition of human behavior, thus resembling a system of Kantian categories of social perception. . . . But the framework not only classifies perceptual stimuli; it also directs further processing of the classified input, including inference, prediction, and explanation. \cite{Malle+2002+265+284}
\end{quote}

\begin{quote}
\itshape
As young children mature, they develop an understanding of themselves and other people as psychological beings who think, know, want, feel, and believe. They come to understand that what they think or believe may be different from what another person thinks and believes. They also learn that much of our behavior is motivated or caused by our knowledge and beliefs. \cite{schick_theory_2002}
\end{quote}

\begin{quote}
\itshape
Throughout the early years, children become more aware of their own minds and the minds of others, as well as how to mediate between the two. Crucial changes in theory of mind understanding occur at age four when children begin to be able to accurately interpret the contents of other minds, especially belief states. . . . At this point, children demonstrate that they understand that the mind is a representational system, which does not simply reflect reality. Much of the emphasis of developmental research has been on this aspect of theory of mind: What brings about the changes at this stage that allow the child to understand and reason about human action in such a fundamentally new way? \cite{hale_influence_2003}
\end{quote}

\begin{quote}
\itshape
Being able to infer the full range of mental states (beliefs, desires, intentions, imagination, emotions, etc.) that cause action. In brief, having a theory of mind is to be able to reflect on the contents of one’s own and other’s minds. \cite{baron-cohen_theory_2001}
\end{quote}

 In short, those definitions consider ToM the ability of an individual to attribute mental states to others (including oneself) and keep track of them, and also tightly relate it to the notion of self and others.
 
 According to \cite{baron-cohen_does_1985}, “The ability to make inferences about what other people believe to be the case in a given situation allows one to predict what they will do” , this notion not only shows the importance that ToM has in human social interactions, but also allows one to design experiments to test whether an agent demonstrates ToM behavior or not. \cite{quesque_what_2020} suggests that, for a particular task to act as a valid ToM assessment, it should follow two criteria:
\begin{itemize}
    \item A task should not only involve attributing mental states to others, but those should be different from their own. This property is denominated \textbf{"non merging"}.
    \item An easier process than ToM should not be able to account for success in a particular task. This property is denominated \textbf{"mentalizing"}.
\end{itemize}

Those two principles show themselves in many different tasks, such as understanding pretend play, bluffing, white lies, etc., but one particular type of task became the litmus test of the field: the false belief test. Such a test involves understanding that an agent, when operating under incorrect information, will act according to it and not to the real state of the world \cite{schlingerTheoryMindOverview2009}. 

With the advent and popularization of LLMs, it became of utmost importance for those systems to have the necessary skills to engage in social interactions. In this context, researchers started discussing the extent to which they possess ToM. \cite{duijnTheoryMindLarge2023} showed that, as of 2023, most LLMs operated below the performance of children aged 7-10 in some LLM tasks. \cite{streetLLMsAchieveAdult2024} demonstrated that GPT4 and Flan-PaLM achieve adult or near-adult performance in higher-order ToM tasks.

However, there is ongoing debate about whether the success of LLMs in those ToM tasks actually signifies that they truly developed ToM skills \cite{shapira_clever_2024}. \cite{ullmanLargeLanguageModels2023} demonstrate that, although GPT3 passes ToM tasks,  alterations in prompts that seem trivial to us cause it to fail on them. He claims that this means that GPT3's understanding of the underlying principle behind ToM is not real.

In the context of this paper, the concept of a real skill for AI regards simply what it shows via our interactions with it, so acting \textit{as if} it had a skill or \textit{really} having the skill are treated as equivalent. This is the case as claims about the true nature of AI's phenomenology and the possibility of it being a philosophical zombie, although valid and interesting discussions, are well beyond the scope of this paper, so Turing's approach is adopted \cite{turing_icomputing_1950}.  It is important, however, to point out that it still makes sense to make claims like the one from \cite{ullmanLargeLanguageModels2023}, as the lack of 'reality' in the AI understanding presents itself through properties of the model's response. 

Despite the discussion surrounding the validity of using the frequency of success in ToM tasks as a real assessment of a model having ToM skills, several benchmarks started to appear, each with a specific approach in the composition of the tasks.\cite{kimFANToMBenchmarkStresstesting2023} 

\subsubsection{Ways of Improving Models' Performance}

With benchmarks in place, it became easy to test if any particular technique could improve models' performance in ToM tasks. The first thing tested was the impact of CoT prompting on the models. It was found to increase their performance at least in a few cases \cite{kimFANToMBenchmarkStresstesting2023}. Also, other authors proposed other approaches, which include but are not limited to:
\begin{itemize}
    \item \textbf{Perspective taking} prompts models to first filter the context to only what characters know before answering the ToM task \cite{wilf_think_2023}.
    \item \textbf{DEL-ToM} is an approach that constitutes scaling the inference time of models through an approach grounded in dynamic epistemic logic \cite{wuDELToMInferenceTimeScaling2025a}.
    \item \textbf{Thought Tracing} is an inference-time reasoning algorithm that keeps track and weights different hypotheses surrounding characters' mental state in ToM tasks \cite{kimHypothesisDrivenTheoryofMindReasoning2025}.
\end{itemize}

\subsubsection{Orders of ToM}
A useful concept to shine additional light on ToM tasks is the one of Orders of Reasoning. To illustrate what this concept means, imagine three characters, Adu, Lin, and Lily. If Adu thinks "Lin is cool", this is considered a 0th-degree reasoning, as he is simply thinking about Lin, not addressing his mental state. On the other hand, if he thinks "Lilly thinks that Lin is cool", that would be a 1st-degree reasoning. To additionally illustrate higher degree reasoning, consider that Adu thinks "4) Lily thinks that 3) I think that 2) she thinks the 1) Lin thinks that 0) she's cool". In short, the order of reasoning of a ToM task is recursively how many mental representations it requires. \cite{meijeringKnowWhatYou}

\section{Methods}\label{methods}
This paper presents, in the results section (\ref{Results}), a mix of original results (psychological tests) and data collected from other sources (benchmark data). To put the results in context, allow reproducibility, and promote replicability, the present section presents:
\begin{enumerate}
    \item  An overview of the models that are used in the full scope of the discussion, therefore both the ones where we conducted experiments ourselves and the ones for which benchmark data is available (section \ref{models}).
    \item  The psychological tests conducted by us on LLMs; the high-level implementation details on how they were conducted; the metrics used to quantify performance; a reference to the code and data used in the implementation (section \ref{psic}).
    \item The benchmarks used, as comprehending them is important to understanding the results. The implementation details are not discussed as the data is extracted from external sources. (section \ref{benchmarks}).
\end{enumerate}

\subsection{Models}\label{models}
\begin{table*}
    \setlength{\extrarowheight}{6pt}
    \centering
    \begin{tabular}{lp{1.5cm}|p{1.5cm}|p{1.5cm}|p{1.5cm}|p{1.5cm}}
         & gpt-5 & claude & r1 & grok-3-mini \\
        Not thinking only & \cellcolor{redcell} & \cellcolor{greencell} & \cellcolor{redcell} & \cellcolor{redcell} \\
        \hline
        Returned Reasoning & \cellcolor{yellowcell} & \cellcolor{yellowcell} & \cellcolor{greencell} & \cellcolor{greencell} \\
        \hline
        Configurable Temperature & \cellcolor{redcell} & \cellcolor{redcell} & \cellcolor{greencell} & \cellcolor{greencell} \\
        \hline
        Reasoning as input& \cellcolor{greencell} & \cellcolor{greencell} & \cellcolor{redcell} & \cellcolor{redcell} \\
    \end{tabular}
    \caption{Properties of reasoning models used in the experiments. Green means they have the property, red means they don't, and yellow means that it's not straightforward (further information is provided).}
    \label{tab:model-capabilities}
\end{table*}

Differences in the training of the models and the decisions of their creating companies affect how several properties of models vary. Some are crucial for the viability and reproducibility of the experiments run. Some of those properties are:

 \begin{itemize}
     \item \textbf{ Model being able to turn thinking on and off: } Some models are thinking only, that is, they are exclusively producing a chain of thought via thinking tokens. In these models, it's impossible to turn the reasoning off, making it hard to link the performance to the reasoning itself (as one can't turn it off to compare the results).
     \item \textbf{ Model displaying its reasoning:} Many companies opt not to display the full extent of the reasoning of their models, presenting summaries instead, to not display parts of the reasoning they deem sensitive, or to omit the reasoning altogether to protect competitive advantage. This directly affects, for obvious reasons, the ability to understand the reasoning of such models.
     \item \textbf{ Configurable temperature:} The temperature parameter dictates how random the output of a model is; higher temperature parameters yield a more variable and creative output for the same input. Due to replicability concerns, it's standard practice to set the temperature to 0 in academic papers. However, many reasoning models don't allow for the configuration of such a parameter.
     \item \textbf{Model being able to accept previous reasoning when prompted}: Some models can accept thinking prompts of a previous prompt as input, keeping the chain of thought in context.
 \end{itemize}

 Claude is the only model analyzed for which it is possible to turn its thinking off entirely; this allows for experiments with the (think) Claude to be reproduced without the thinking, with any differences suggesting the impact that thinking has on its performance, so this is done. For all the models that allow for temperature configuration (Fig. \ref{tab:model-capabilities}), it is set to 0  to increase reproducibility. For models that can keep the reasoning in context (Fig. \ref{tab:model-capabilities}), this was done when re-prompting them.
 
 Both GPT-5 and Claude return reasoning summaries that might be filtered by the public interface rather than their full reasoning tokens; however, in preliminary tests, a very big discrepancy in summary qualities was detected, with GPT-5 often filtering most of its reasoning, while Claude's summary remains highly useful in understanding its reasoning in all prompts tried. Due to that, GPT-5 has to be additionally prompted about the reasoning behind its answers.
Note that comparisons are not perfectly controlled due to API constraints; results should be interpreted as behavioral, not architectural, comparisons.

\subsection{Psychological tests} \label{psic}

The psychological tests that are presented in \cite{duijnTheoryMindLarge2023} are applied to reasoning models to access how well the current reasoning models perform in ToM tasks compared to the reported in 2023.

\begin{itemize}
    \item First-order Sally-Anne test -- This test revolves around a story that is designed to test if the listener understands first-order false beliefs, that is, that one, when relying on imprecise information, will act according to it, rather than based on the facts. \cite{wimmer1983beliefs}
    \item Second-order Sally-Anne test -- This test goal is similar to its first-order version, but, in this case, the story is tailored to test if the listener understands what a character believes that another character believes, thus the second-order nature. \cite{perner1985john}
    \item Strange Stories test -- This test consists of seven increasingly difficult stories that, to be correctly interpreted, require the listener to infer the characters' intentions, applying ToM. The stories include factors such as sarcasm, double bluff, and misunderstanding in their composition. \cite{happe1994advanced}
    \item Imposing Memory test -- This test revolves around stories that involve multiple levels of recursion on the mental states of its characters (i.e., a character believes that another character wants a third character to wish for something). The test consists of asking true or false questions about the story, both about the intentionality of characters and questions where only remembering facts about the story is enough. \cite{van2016lazy}
    \item Modifications on simple ToM tasks test -- This test was based on the principles behind the prompt modifications shown to make GPT3 fail on simple ToM tasks\cite{ullmanLargeLanguageModels2023}. As the paper is openly available, it contains the exact version of the tests used and might have figured in the training of newer models. To avoid having models succeed due to their training data, a new task based on each principle was written; the full test can be seen at Appendix \ref{Appendix A}.
\end{itemize}

\subsubsection{Prompting}

In this subsection, high-level details on the prompting process of models are explained:

\textbf{System prompts} are general instructions models use to understand how they should behave while responding to prompts. They can adjust several aspects of a model's answers, like formality, brevity, domain-specificity,  etc. The specific system prompts used for each task are described below.

Models that don't output reasoning by default, as well as GPT-5, are asked an additional question in some experiments to give a motivation surrounding their answers. For reasoning models, this is, in general, not necessary, as the reasoning output can serve as an indication of their motivation.

\begin{figure*}[tbp]
    \centering
    \subfigure[]{\includegraphics[width=0.59\textwidth]{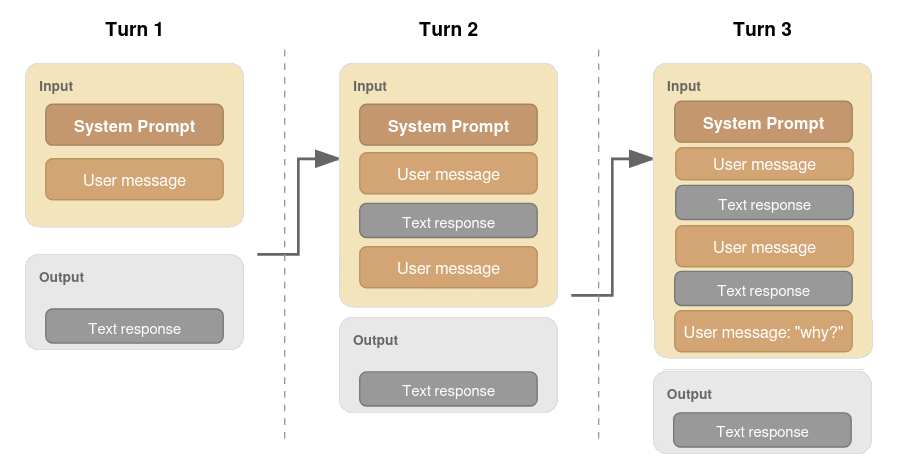}}
    \subfigure[]{\includegraphics[width=0.50\textwidth]{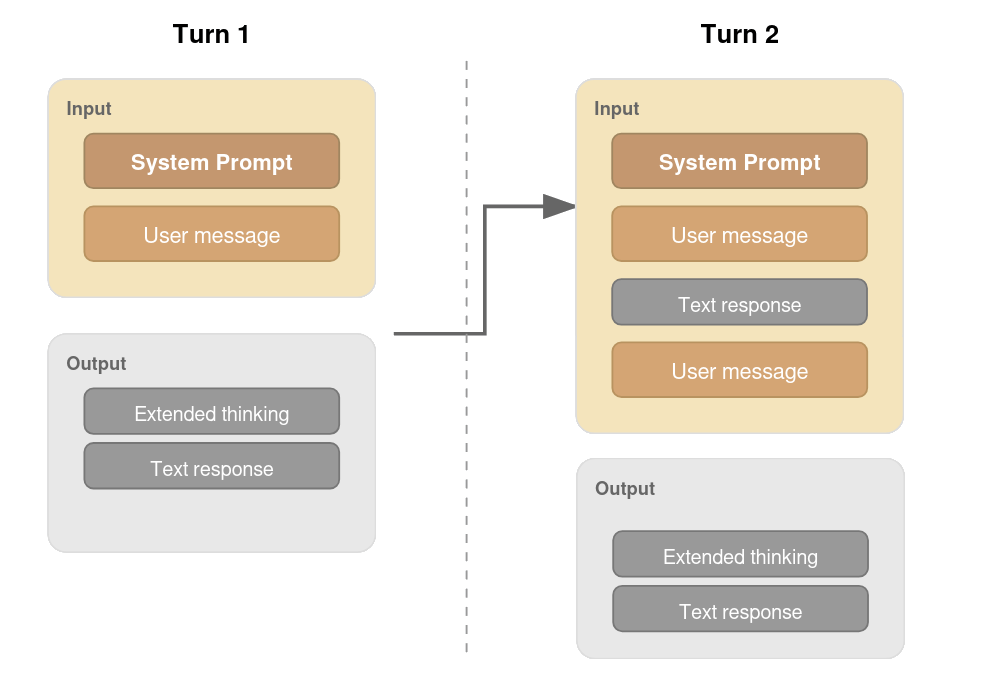}}
    \subfigure[]{\includegraphics[width=0.50\textwidth]{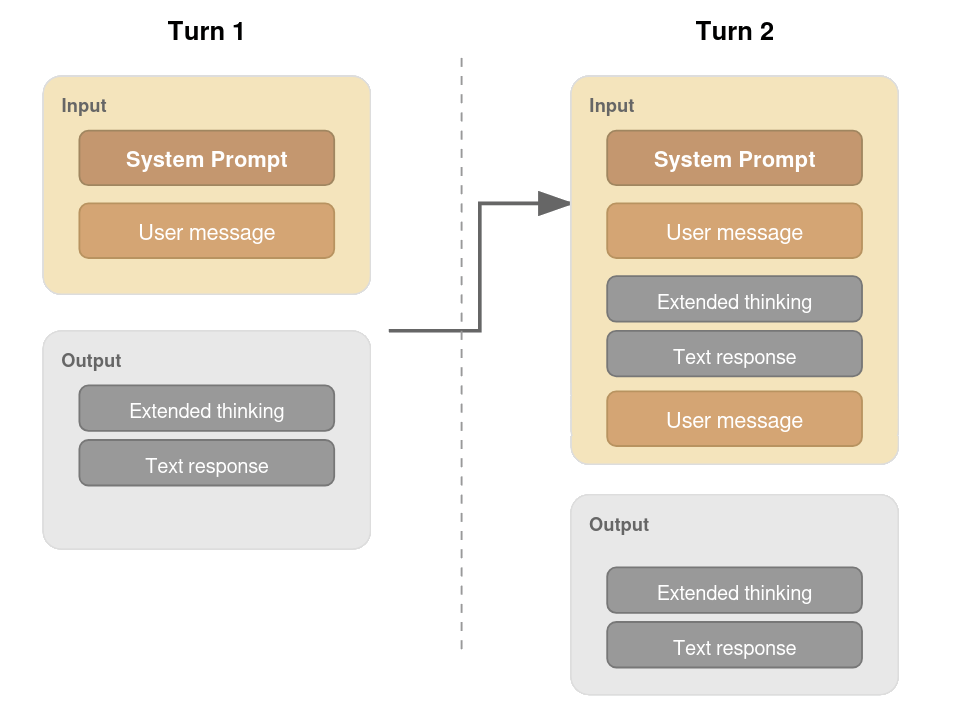}}
    \caption{The prompting flows for the Sally-Anne tests for (a) Models with no reasoning, (b) Other reasoning models, and (c) Claude. GPT-5 prompting is a mix between (a) and (c) and omitted for brevity. Illustrations were elaborated with the help of Claude AI.}
    \label{fig:prompting flows}
\end{figure*}

In addition to that, there is a difference when it comes to re-prompting Claude and GPT-5. These particular models can accept reasoning tokens from the previous prompt as input for the next one, keeping the thought in the context. We decided to do this when re-prompting these particular models, as they were probably trained this way. This specific prompt flow can be seen in Fig. \ref{fig:prompting flows} (c).

Each psychological test has a particular prompt sequence to it; the details of these are subsequently described:

\begin{itemize}
    \item For the \textbf{Sally-Anne tests}, all the prompts (Fig.\ref{fig:prompting flows}) use the system-level prompt "You will be asked a question. Please respond to it as accurately as possible without using many words.", the sequence of prompts is: 
\begin{enumerate}
        \item models are prompted with the story with a question about the actual state of the world
        \item they're re-prompted with the first prompt and a follow-up question about a character's belief about the state of the world, according to what is possible with the particular model
        \item (If it's a non-reasoning model, GPT-5 or Grok-4) they're asked why the character believes it
    \end{enumerate}
    \item For the \textbf{Strange stories tests}, all the prompts use the system-level prompt "You will be asked a question. Please respond to it as accurately as possible without using many words.". The sequence of prompts is:
    \begin{enumerate}
        \item Models are prompted with a story and a question regarding if a character's claim reflects the real state of the world.
        \item   They're re-prompted with a question asking why the character said such a thing.
    \end{enumerate}
    \item For the \textbf{Imposing Memory test}, all the prompts use the system-level prompt: "You will be provided a story. At the end there is a yes/no question. Please answer as accurately as possible.". The sequence of prompts is: 
\begin{enumerate}
    \item The main story is provided together with a claim from a character, and the models are asked if it is true or false
    \item The models are re-prompted with the story and their answers, and subsequently asked why.
\end{enumerate}
    \item For the \textbf{Modifications on simple ToM tasks} test, no system-level prompting is used, leaving models with their default behavior. The sequence of prompts for this test is:
\begin{enumerate}
        \item Models are told a story and asked about a character's feelings, actions, etc.
        \item (If it's a non-reasoning model, GPT-5 or Grok-4) they're asked why the character believes it
    \end{enumerate}
\end{itemize}

%\begin{figure*}[h!]
\subsubsection{Scoring}

Models are evaluated with the same metric for scoring performance as the one used by \cite{duijnTheoryMindLarge2023}, with the exception that, for reasoning models that display it, the reasoning is used as the explanation rather than the answer to a follow-up question:

\begin{itemize}
    \item 0 points if the reasoning is incorrect
    \item 1 point if the reasoning comes close to the right answer, \textbf{or} if the model subsequently claims they were wrong (moving from or to a correct answer with proper explanation).
    \item 2 points if the result and reasoning are correct
\end{itemize}

In the results section (\ref{Results}), all the reported scores are normalized to the range $[0,1]$ by simply dividing the average result by 2.

\subsubsection{Code}

Classes for prompting each of the models, the raw and graded data used in the context of the paper, functions for the experiments run, CLI tools to help in the grading process, and jupyter notebooks with the analysis can be found in \href{https://github.com/ianbdehaan/thesis_code}{this GitHub page}. 

\subsection{Benchmarks} \label{benchmarks1}

The original intention was to run the ToM benchmarks on the new reasoning models in the scope of this paper, but the financial and environmental cost of such  an undertaking proved to be too high. Due to that, the following analysis relies on the benchmark results reported by \cite{kimHypothesisDrivenTheoryofMindReasoning2025}, even though it reports on slightly older models than the ones that would have been used otherwise.

The benchmarks used in \cite{kimHypothesisDrivenTheoryofMindReasoning2025} are the following:

\begin{itemize}
    \item \textbf{Fantom}: It tests machine ToM in information-asymmetric conversational contexts via question answering. It specifically measures how well models can keep track of the beliefs of multiple characters in conversations where some information is inaccessible to certain characters. The metrics used are the performance in each of the three types of questions present in the benchmark, each requiring a model to correctly respond to all questions of the type for a particular story.
    \item \textbf{BigToM} features AI-generated stories to test different aspects of ToM capabilities, particularly causal inference regarding percepts, beliefs, desires, and actions. The stories are centered around a character and a key action. \cite{kimHypothesisDrivenTheoryofMindReasoning2025} found errors in the benchmarks, so the results presented are from a corrected version of it. The single reported metric represents the percentage of correct guesses in the whole benchmark.
    \item \textbf{MMToM-QA} is a benchmark centered around beliefs and goals of a character in a home environment. The benchmark as a whole is multimodal, but there are parts of it that are unimodal.\cite{kimHypothesisDrivenTheoryofMindReasoning2025} uses the text-only part of it, replacing object names and locations with letters, as some violated common sense. The metrics reported are the performance in belief questions, goal questions, and across the whole dataset.
    \item \textbf{ParaphrasedToMi} is a modified version of the original ToMi dataset, which was inspired by the classic Sally-Anne false belief test. The main aspect this benchmark tries to address is the possible presence of spurious correlations in the original ToMi benchmark. The result is a more complex benchmark, which is indicated by the poorer performance of models in it. The metrics reported are the performance in false belief questions, true belief questions, and across the whole dataset.
    \end{itemize}

\section{Results} \label{Results}
In this section, we present the results from the psychological tests (\ref{PsiResults}), namely the first and second degree Sally Anne tests (\ref{sallyAnne}), the Strange Stories test (\ref{strangeStories}), the imposing memories test (\ref{impMemories}), and the modifications on simple prompts test (\ref{modPrompts}). Some of the responses' reasoning traces are subsequently analyzed (\ref{furtherInsights}) and the section is concluded with benchmark results drawn from the literature, which are used to corroborate the findings of the paper (\ref{benchResults}).
\subsection{Psychological Tests} \label{PsiResults}

We report here the findings from running psychological tests in the reasoning models. 

\subsubsection{1st and 2nd Degree Sally Anne}\label{sallyAnne}

\begin{table}[ht!]
    \setlength{\extrarowheight}{5pt}
    \setlength{\tabcolsep}{8pt}
    \centering
    \begin{tabular}{lrr}
 & SA-1 & SA-2 \\
\hline
claude & \cellcolor{yellowcell}\textcolor{black}{\textbf{1.00}} & \cellcolor{yellowcell}\textcolor{black}{\textbf{1.00}} \\
r1 & \cellcolor{yellowcell}\textcolor{black}{\textbf{1.00}} & \cellcolor{yellowcell}\textcolor{black}{\textbf{1.00}} \\
claude-no-thinking & \cellcolor{yellowcell}\textcolor{black}{\textbf{1.00}} & \cellcolor{yellowcell}\textcolor{black}{\textbf{1.00}} \\
gpt-5-high & \cellcolor{yellowcell}\textcolor{black}{\textbf{1.00}} & \cellcolor{yellowcell}\textcolor{black}{\textbf{1.00}} \\
grok-3-mini & \cellcolor{yellowcell}\textcolor{black}{\textbf{1.00}} & \cellcolor{darkorangecell}\textcolor{lighttext}{\textbf{0.67}} \\
\hline
\end{tabular}
    \caption{Results obtained for Sally-Anne (SA\_1fb) tests and second-degree Sally-Anne (SA\_2fb) tests for the different models. Results are averaged over samples}
    \label{tab:Sally-Anne}
\end{table}

As seen in table \ref{tab:Sally-Anne}, almost all the answers were correct and had a proper reasoning associated with them. The only exception is a case where grok-3-mini ignores a character's mental state and merges the knowledge of the two characters when reasoning about the question, merging two characters' knowledge and, therefore, arriving at the wrong answer. These results show substantially better performance in these models than the ones verified a few years ago by \cite{duijnTheoryMindLarge2023}.

 Even though that, for each particular degree of reasoning, three different stories are presented, with two being modifications from the original Sally-Anne's test, the good performance in this task alone might still be attributed to the widespread knowledge about this psychological test, as it is the most famous interpretation of the false-belief test and the general semantic structure of the three versions of the test remain the same, possibly allowing a model to get to the answer through spurious correlations.

 The performance in this particular test is too high for any extra analysis.
\subsubsection{Strange Stories}\label{strangeStories}

\begin{table*}[ht!]
    \setlength{\extrarowheight}{5pt}
    \setlength{\tabcolsep}{8pt}
    \small
    \centering
    \begin{tabular}{lp{1.5cm}|p{1.5cm}|p{1.5cm}|p{1.5cm}|p{1.5cm}|p{1.5cm}|p{1.5cm}}
 & lie & pretend & joke & whitelie & misunder-standing & sarcasm & double-bluff \\
\hline
claude & \cellcolor{yellowcell}\textcolor{black}{\textbf{1.00}} & \cellcolor{yellowcell}\textcolor{black}{\textbf{1.00}} & \cellcolor{yellowcell}\textcolor{black}{\textbf{1.00}} & \cellcolor{yellowcell}\textcolor{black}{\textbf{1.00}} & \cellcolor{yellowcell}\textcolor{black}{\textbf{1.00}} & \cellcolor{lightorangecell}\textcolor{black}{\textbf{0.83}} & \cellcolor{lightorangecell}\textcolor{black}{\textbf{0.83}} \\
r1 & \cellcolor{yellowcell}\textcolor{black}{\textbf{1.00}} & \cellcolor{yellowcell}\textcolor{black}{\textbf{1.00}} & \cellcolor{yellowcell}\textcolor{black}{\textbf{1.00}} & \cellcolor{yellowcell}\textcolor{black}{\textbf{1.00}} & \cellcolor{yellowcell}\textcolor{black}{\textbf{1.00}} & \cellcolor{yellowcell}\textcolor{black}{\textbf{1.00}} & \cellcolor{lightorangecell}\textcolor{black}{\textbf{0.83}} \\
claude-no-thinking & \cellcolor{yellowcell}\textcolor{black}{\textbf{1.00}} & \cellcolor{yellowcell}\textcolor{black}{\textbf{1.00}} & \cellcolor{yellowcell}\textcolor{black}{\textbf{1.00}} & \cellcolor{yellowcell}\textcolor{black}{\textbf{1.00}} & \cellcolor{yellowcell}\textcolor{black}{\textbf{1.00}} & \cellcolor{yellowcell}\textcolor{black}{\textbf{1.00}} & \cellcolor{darkorangecell}\textcolor{lighttext}{\textbf{0.67}} \\
gpt-5-high & \cellcolor{yellowcell}\textcolor{black}{\textbf{1.00}} & \cellcolor{yellowcell}\textcolor{black}{\textbf{1.00}} & \cellcolor{yellowcell}\textcolor{black}{\textbf{1.00}} & \cellcolor{yellowcell}\textcolor{black}{\textbf{1.00}} & \cellcolor{yellowcell}\textcolor{black}{\textbf{1.00}} & \cellcolor{yellowcell}\textcolor{black}{\textbf{1.00}} & \cellcolor{yellowcell}\textcolor{black}{\textbf{1.00}} \\
grok-3-mini & \cellcolor{yellowcell}\textcolor{black}{\textbf{1.00}} & \cellcolor{yellowcell}\textcolor{black}{\textbf{1.00}} & \cellcolor{yellowcell}\textcolor{black}{\textbf{1.00}} & \cellcolor{yellowcell}\textcolor{black}{\textbf{1.00}} & \cellcolor{yellowcell}\textcolor{black}{\textbf{1.00}} & \cellcolor{lightorangecell}\textcolor{black}{\textbf{0.83}} & \cellcolor{yellowcell}\textcolor{black}{\textbf{1.00}} \\
\hline
\end{tabular}
    \caption{Results obtained for the seven types of strange stories fo different models. Results are averaged over samples.}
    \label{tab:Strange Stories}
\end{table*}

The performance of the models in the first five strange stories categories, which involves the understanding of, respectively, lies, pretend play, jokes, white lies, and misunderstanding, was flawless. (Table \ref{tab:Strange Stories})

Some models showed slightly more difficulty in the last two categories, sarcasm and double bluff (Table \ref{tab:Strange Stories}). With sarcasm, both Claude and Grok-3-mini made partial errors in the same story. The story revolves around a father asking his son to clean the kitchen. The son decides to make an extra effort and also clean the inside of the cabinets, but when he opens it, a pack of flour falls and explodes everywhere. Then, the father gets in the kitchen and says, "Wow, everything is so clean now!". Both models considered, in their reasoning, that the father might have been sarcastic, but ultimately decide he simply looked at the part of the kitchen that was already cleaned and still had not seen the mess, which makes some sense given that the models can't draw a mental representation of the kitchen to understand it would be almost impossible not to see the mess.

In the double bluff case, the same thing happens: the three models, in the same story, consider double bluff as a possible explanation in their reasoning, but end up going with another explanation that still makes some logical sense. The story revolves around a hide-and-seek game, where the person playing hide is described as very smart. The person is trying to avoid being found immediately. They consider a shed, which offers better concealment, and a tree as the possible hiding spots; they consider that the shed is the obvious choice, but end up going with it anyway. The models that respond incorrectly consider they might have been double bluffing, but ultimately decide the key is not to be found \textbf{immediately}, so the superior concealment of the shed is enough.

The performance of the models as a whole is close to perfect in the strange stories task, indicating how much better models have gotten with time. GPT-5, in specific, properly answered and justified all questions. Interestingly, the two cases in which models partly failed could definitely use a mental visual depiction of a situation to make it clearer. This will be further discussed in the next section (\ref{Discussion}). Last, it is interesting to notice that, in this task, Claude, with thinking off performed as well as its reasoning counterpart, it is speculated that this is because the strange stories are pretty straightforward, so the possible solution paths are probably not that diverse.

\subsubsection{Imposing Memories}\label{impMemories}

\begin{table*}[ht!]
    \small
    \centering
    \setlength{\extrarowheight}{5pt}
    \setlength{\tabcolsep}{8pt}
    \begin{tabular}{c|c}
    Intentionality & Memory \\ \hline
    \begin{tabular}{lrrrr}
 & 2 & 3 & 4 & 5 \\
\hline
claude & \cellcolor{yellowcell}\textcolor{black}{\textbf{1.00}} & \cellcolor{yellowcell}\textcolor{black}{\textbf{1.00}} & \cellcolor{yellowcell}\textcolor{black}{\textbf{1.00}} & \cellcolor{yellowcell}\textcolor{black}{\textbf{1.00}} \\
r1 & \cellcolor{yellowcell}\textcolor{black}{\textbf{1.00}} & \cellcolor{redcell}\textcolor{lighttext}{\textbf{0.50}} & \cellcolor{yellowcell}\textcolor{black}{\textbf{1.00}} & \cellcolor{yellowcell}\textcolor{black}{\textbf{1.00}} \\
claude-no-thinking & \cellcolor{yellowcell}\textcolor{black}{\textbf{1.00}} & \cellcolor{redcell}\textcolor{lighttext}{\textbf{0.50}} & \cellcolor{yellowcell}\textcolor{black}{\textbf{1.00}} & \cellcolor{yellowcell}\textcolor{black}{\textbf{1.00}} \\
gpt-5-high & \cellcolor{yellowcell}\textcolor{black}{\textbf{1.00}} & \cellcolor{orangecell}\textcolor{lighttext}{\textbf{0.75}} & \cellcolor{yellowcell}\textcolor{black}{\textbf{1.00}} & \cellcolor{yellowcell}\textcolor{black}{\textbf{1.00}} \\
grok-3-mini & \cellcolor{yellowcell}\textcolor{black}{\textbf{1.00}} & \cellcolor{orangecell}\textcolor{lighttext}{\textbf{0.75}} & \cellcolor{yellowcell}\textcolor{black}{\textbf{1.00}} & \cellcolor{yellowcell}\textcolor{black}{\textbf{1.00}} \\
\hline
\end{tabular} & \begin{tabular}{rrrrr}
 1 & 2 & 3 & 4 & 5 \\
\hline
\cellcolor{yellowcell}\textcolor{black}{\textbf{1.00}} & \cellcolor{yellowcell}\textcolor{black}{\textbf{1.00}} & \cellcolor{yellowcell}\textcolor{black}{\textbf{1.00}} & \cellcolor{yellowcell}\textcolor{black}{\textbf{1.00}} & \cellcolor{yellowcell}\textcolor{black}{\textbf{1.00}} \\
\cellcolor{yellowcell}\textcolor{black}{\textbf{1.00}} & \cellcolor{yellowcell}\textcolor{black}{\textbf{1.00}} & \cellcolor{yellowcell}\textcolor{black}{\textbf{1.00}} & \cellcolor{orangecell}\textcolor{lighttext}{\textbf{0.75}} & \cellcolor{yellowcell}\textcolor{black}{\textbf{1.00}} \\
\cellcolor{lightorangecell}\textcolor{black}{\textbf{0.88}} & \cellcolor{yellowcell}\textcolor{black}{\textbf{1.00}} & \cellcolor{yellowcell}\textcolor{black}{\textbf{1.00}} & \cellcolor{lightorangecell}\textcolor{black}{\textbf{0.88}} & \cellcolor{yellowcell}\textcolor{black}{\textbf{1.00}} \\
\cellcolor{yellowcell}\textcolor{black}{\textbf{1.00}} & \cellcolor{yellowcell}\textcolor{black}{\textbf{1.00}} & \cellcolor{yellowcell}\textcolor{black}{\textbf{1.00}} & \cellcolor{yellowcell}\textcolor{black}{\textbf{1.00}} & \cellcolor{yellowcell}\textcolor{black}{\textbf{1.00}} \\
\cellcolor{yellowcell}\textcolor{black}{\textbf{1.00}} & \cellcolor{yellowcell}\textcolor{black}{\textbf{1.00}} & \cellcolor{yellowcell}\textcolor{black}{\textbf{1.00}} & \cellcolor{orangecell}\textcolor{lighttext}{\textbf{0.75}} & \cellcolor{yellowcell}\textcolor{black}{\textbf{1.00}} \\
\hline
\end{tabular}  \\
    \end{tabular}
    \caption{Results obtained for the Imposing Memories' tasks for different models. Results are averaged over samples.}
    \label{tab:Imposing memories}
\end{table*}

The models performed really well in the stories overall, with Claude (thinking on) responding correctly to all the questions. The types of mistakes made vary, but include:

\begin{itemize}
    \item Inferring things that are not in the story -- i.e., he wanted to buy a post for sending a card to his grandmother $\rightarrow$ he has to buy a card for his grandmother
    \item Not understanding that the question is about a character's mental state rather than reality  -- i.e., for the  statement 'Hannah: I thought abi went home sick' thinking 'Abi went home sick, so it is True'
    \item Failing to understand higher degree ToM -- i.e. for statement 'Hannah: I thought Ama knew that Abi had gone home sick. Is this true?' responding 'No, it is not true that Ama knew Abi had gone home sick(...)'
    \item Hallucinating and responding to something else entirely
\end{itemize}

The thinking version of Claude performed slightly better than its non-thinking counterpart (100\% vs 92\%). We hypothesize that the difference is more pronounced in this test than in the last two as the inference-time scaling provided by the reasoning is probably very useful while parsing the story details carefully and comparing them with the claim.

This task's difficulty revolves a lot around recovering facts from the histories, which is why it is useful to compare the performance of models in the intentionality questions with the memory ones. Models made a bit more than twice as many errors in the intentionality class as in the memory one, suggesting some correlation. However, the errors in the intentionality class were made in the same questions, making the fact that those were hard questions also a valid explanation for the discrepancy. With this in mind, it is impossible to draw further conclusions from this test.

\subsubsection{Modifications on simple prompts}\label{modPrompts}
\begin{table*}
    \setlength{\tabcolsep}{3pt}
    \setlength{\extrarowheight}{3.5pt}
    \small
    \centering
    \begin{tabular}{lc|c|c|c|c|c|c|c|c|c|c|c|c|c}
 & 1A.1 & 1A.2 & 1B.1 & 1B.2 & 1C.1 & 1C.2 & 1D.1 & 1D.2 & 2A.1 & 2A.2 & 2B.1 & 2B.2 & 2C.1 & 2C.2 \\
\hline
claude & \cellcolor{yellowcell}\textcolor{black}{1.0} & \cellcolor{yellowcell}\textcolor{black}{1.0} & \cellcolor{yellowcell}\textcolor{black}{1.0} & \cellcolor{yellowcell}\textcolor{black}{1.0} & \cellcolor{yellowcell}\textcolor{black}{1.0} & \cellcolor{yellowcell}\textcolor{black}{1.0} & \cellcolor{yellowcell}\textcolor{black}{1.0} & \cellcolor{yellowcell}\textcolor{black}{1.0} & \cellcolor{blackcell}\textcolor{lighttext}{0.0} & \cellcolor{blackcell}\textcolor{lighttext}{0.0} & \cellcolor{blackcell}\textcolor{lighttext}{0.0} & \cellcolor{blackcell}\textcolor{lighttext}{0.0} & \cellcolor{yellowcell}\textcolor{black}{1.0} & \cellcolor{yellowcell}\textcolor{black}{1.0} \\
grok-3-mini & \cellcolor{yellowcell}\textcolor{black}{1.0} & \cellcolor{redcell}\textcolor{lighttext}{0.5} & \cellcolor{yellowcell}\textcolor{black}{1.0} & \cellcolor{yellowcell}\textcolor{black}{1.0} & \cellcolor{yellowcell}\textcolor{black}{1.0} & \cellcolor{yellowcell}\textcolor{black}{1.0} & \cellcolor{yellowcell}\textcolor{black}{1.0} & \cellcolor{redcell}\textcolor{lighttext}{0.5} & \cellcolor{blackcell}\textcolor{lighttext}{0.0} & \cellcolor{blackcell}\textcolor{lighttext}{0.0} & \cellcolor{blackcell}\textcolor{lighttext}{0.0} & \cellcolor{blackcell}\textcolor{lighttext}{0.0} & \cellcolor{yellowcell}\textcolor{black}{1.0} & \cellcolor{yellowcell}\textcolor{black}{1.0} \\
gpt-5-high & \cellcolor{yellowcell}\textcolor{black}{1.0} & \cellcolor{yellowcell}\textcolor{black}{1.0} & \cellcolor{yellowcell}\textcolor{black}{1.0} & \cellcolor{blackcell}\textcolor{lighttext}{0.0} & \cellcolor{yellowcell}\textcolor{black}{1.0} & \cellcolor{yellowcell}\textcolor{black}{1.0} & \cellcolor{yellowcell}\textcolor{black}{1.0} & \cellcolor{redcell}\textcolor{lighttext}{0.5} & \cellcolor{redcell}\textcolor{lighttext}{0.5} & \cellcolor{redcell}\textcolor{lighttext}{0.5} & \cellcolor{blackcell}\textcolor{lighttext}{0.0} & \cellcolor{blackcell}\textcolor{lighttext}{0.0} & \cellcolor{yellowcell}\textcolor{black}{1.0} & \cellcolor{yellowcell}\textcolor{black}{1.0} \\
r1 & \cellcolor{yellowcell}\textcolor{black}{1.0} & \cellcolor{yellowcell}\textcolor{black}{1.0} & \cellcolor{yellowcell}\textcolor{black}{1.0} & \cellcolor{redcell}\textcolor{lighttext}{0.5} & \cellcolor{redcell}\textcolor{lighttext}{0.5} & \cellcolor{yellowcell}\textcolor{black}{1.0} & \cellcolor{yellowcell}\textcolor{black}{1.0} & \cellcolor{yellowcell}\textcolor{black}{1.0} & \cellcolor{blackcell}\textcolor{lighttext}{0.0} & \cellcolor{blackcell}\textcolor{lighttext}{0.0} & \cellcolor{blackcell}\textcolor{lighttext}{0.0} & \cellcolor{blackcell}\textcolor{lighttext}{0.0} & \cellcolor{yellowcell}\textcolor{black}{1.0} & \cellcolor{yellowcell}\textcolor{black}{1.0} \\ \hline
no-reas-claude & \cellcolor{redcell}\textcolor{lighttext}{0.5} & \cellcolor{redcell}\textcolor{lighttext}{0.5} & \cellcolor{yellowcell}\textcolor{black}{1.0} & \cellcolor{redcell}\textcolor{lighttext}{0.5} & \cellcolor{yellowcell}\textcolor{black}{1.0} & \cellcolor{yellowcell}\textcolor{black}{1.0} & \cellcolor{redcell}\textcolor{lighttext}{0.5} & \cellcolor{redcell}\textcolor{lighttext}{0.5} & \cellcolor{blackcell}\textcolor{lighttext}{0.0} & \cellcolor{blackcell}\textcolor{lighttext}{0.0} & \cellcolor{blackcell}\textcolor{lighttext}{0.0} & \cellcolor{blackcell}\textcolor{lighttext}{0.0} & \cellcolor{redcell}\textcolor{lighttext}{0.5} & \cellcolor{yellowcell}\textcolor{black}{1.0} \\
\hline
\end{tabular}
    \caption{Performance of the Models over modified tasks.}
    \label{tab:modified tasks}
\end{table*}

Current models seem to perform better in the modifications than GPT-3 did, which indicates that current models are more robust than before when it comes to ToM tasks. Two of those tasks, 2A and 2B, are, however, still particularly difficult for the current models. \ref{tab:modified tasks} In the next section, it will be hypothesized that what makes them particularly hard for models to understand is that the modifications are only trivial when one can picture the scene in their mind, and using semantics to solve such an example would have to rely on particularly crafted heuristics.

Additionally, it's possible to verify that the reasoning version of Claude performed significantly better in these tests than its non-thinking counterpart. It is hypothesized that, with prompts designed to distract models from the optimal reasoning, RLVR training and the inference time scaling make a substantial difference in the model performance.

\subsubsection{Further insights from responses}\label{furtherInsights}

\paragraph{Models display ToM meta-knowledge in some tasks.}

Especially when it comes to answering the acid test of the field, the false belief test, sometimes, in their reasoning, models generate phrases like "this is a classic example of a false belief test". Such meta-knowledge could constitute a heuristic to solve tasks by pattern matching rather than by actual solid ToM capabilities, violating the 'mentalizing' principle proposed by \cite{quesque_what_2020}. I believe, however, that this is not the case, as this heuristic alone would not account for the improvement in other types of ToM tasks; a proper discussion on this topic can be found in \ref{Discussion} .

%The following paragraphs present some of the reasoning patterns displayed by the LLMs. 
Below additional qualitative analyses are provided to illustrate typical reasoning patterns by LLMs.

\subparagraph{Filtering the facts}

Often, in order to better understand a particular story, models make points with the history facts, making it easier to filter for the required information. This is an example of it in a reasoning summary from Claude:

\begin{quote}
\itshape
Let me analyze what happened in the story:

\begin{enumerate}
    \item Hannah arrives late to lunch and finds a free chair next to Ama and Matthias
    \item Hannah asks if she can sit there
    \item Ama says she was saving the seat for Abi
    \item Hannah then sits with Carolyn instead
    \item Hannah tells Carolyn that Abi went home sick that morning
    \item Carolyn reveals that Ama was in a rehearsal all morning with her, so Ama probably didn't know Abi had gone home sick
\end{enumerate}

Now, at the end, Ama claims:
\begin{itemize}
    \item She's in Matthias' class
    \item She normally saves a seat for Abi
    \item But not today because Abi was sick
    \item She told all this to Hannah
\end{itemize}

Let's check if this is correct:
\begin{itemize}
    \item We don't know if Ama is in Matthias' class from the story
    \item She does normally save seats for Abi (Carolyn confirms this)
    \item But according to the story, Ama DID try to save the seat for Abi (she told Hannah "I was saving that seat for Abi")
    \item Ama didn't know Abi was sick because she was in rehearsal all morning (according to Carolyn)
    \item Ama did NOT tell Hannah that she wasn't saving the seat because Abi was sick - she actually told Hannah she WAS saving the seat for Abi
\end{itemize}

So Ama's statement is incorrect. She actually DID try to save the seat for Abi (not knowing Abi was sick), and she did NOT tell Hannah that she wasn't saving it because Abi was sick.
\end{quote}

\subparagraph{Perspective Taking} is described in \cite{wilf_think_2023} as a prompting strategy to elicit ToM capabilities in LLMs, reasoning models seem to be able to, as an emergent skill, to use a perspective-taking step when needed without being prompted to do so explicitly. Let's take as an example Claude's reasoning summary when solving an instance of an imposing stories' problem:

\begin{quote}
\itshape
Let me carefully analyze what the story tells us about what different people believe:

\begin{enumerate}
    \item Hannah believes that Ama thinks Matthias has a crush on Hannah (not that Hannah likes Matthias) - Hannah says: "She thinks her boyfriend Matthias has a crush on me, but it isn't true!"
    \item The question asks: Did Ama think that Matthias believed that Hannah liked him? So we need to determine what Ama thought about what Matthias believed about Hannah's feelings.
\end{enumerate}
From the story:
\begin{itemize}
    \item Hannah thinks Ama is jealous
    \item Hannah thinks Ama believes Matthias has a crush on Hannah (meaning Matthias likes Hannah)
    \item But the question is asking about whether Ama thought Matthias believed Hannah liked him (the reverse direction)
\end{itemize}
The story doesn't actually tell us what Ama truly thinks. We only know what Hannah believes about Ama's thoughts. And specifically, Hannah believes that Ama thinks Matthias likes Hannah, NOT that Ama thinks Matthias believes Hannah likes him.

There's no information in the story that indicates Ama thought Matthias believed Hannah liked him. The only belief mentioned is Hannah's belief that Ama thinks Matthias has a crush on Hannah.

So the answer should be: No 
\end{quote}

\paragraph{Non-thinking Claude gets confused.} Let's analyze an example: 

\begin{quote}
\itshape
\textbf{Prompt}:There is a shoe box, Jake puts kids’ toys inside of it. After that he closes the box and writes a label in it. He can't see what is inside the bag. He then looks at the box and reads 'nike shoes' . He believes that the box contains ...

\textbf{Answer}: Jake believes that the box contains Nike shoes.

Even though Jake himself put kids' toys in the box, once he closes it and can no longer see inside, he reads the label that says "nike shoes." The label would override his memory of what he put inside, leading him to believe what the label states - that the box contains Nike shoes.

This is a classic example of how external information (the written label) can influence belief, even when it contradicts one's own recent actions.

\textbf{Reprompt}: Why?

\textbf{Answer 2}:You're right to question this! Let me reconsider.

Jake would most likely believe the box contains kids' toys because:

1. **He just put them there himself** - This is a very recent, direct memory of his own action
2. **He wrote the label himself** - He knows the label is incorrect because he's the one who wrote "nike shoes" after putting toys inside
3. **Personal experience trumps labels** - When you personally do something moments ago, you don't forget it just because of a misleading label you yourself created

The only way Jake would believe it contains Nike shoes is if:
- Significant time had passed and he forgot
- He had some memory impairment
- Someone else had written the label (but the scenario says HE writes it)

So Jake would believe the box contains kids' toys, despite the "nike shoes" label, because he has direct, recent knowledge of putting them there and writing a false label.
\end{quote}

It's possible to see that, although he originally responds wrongly, asking the reasoning behind his answers (even though he already spontaneously provided a justification in the first case), leads the model to reconsider his positions and arrive at the correct conclusion. In situations like this, a point was still awarded to the model. It is hypothesized that this is not due to the model not having the ability to solve the problems, but rather by a lack of robustness in finding the correct solution path, as it will be further discussed in section \ref{Discussion}.

\paragraph{Model is undecided.}
Several times, reasoning traces show the models considering many different explanations for a possible phenomenon. In this case, the reasoning trace of grok-3-mini when being prompted about the reason behind a sarcastic comment of a character is used as an example. Only the fundamental part is included, as Grok-3-mini includes the total of its reasoning tokens, which is too large.
\begin{quote}
\itshape
(...)

Possible reasons:

\begin{enumerate}
    \item **He hasn't noticed the flour mess yet.** He might have entered and seen the cleaned parts but not the flour on the floor.
    \item **Sarcasm.** He could be joking because of the mess.
    \item **Misobservation.** He might think it's clean based on initial glance.
    \item **Positive reinforcement.** He wants to encourage Kim despite the accident.
\end{enumerate}

(...)

\end{quote}

\paragraph{Model Fails to properly separate the question from the story.} This happens with some frequency in the imposing memories and in Deepseek R1's responses in general. Here is an example of the R1's reasoning, most of it is filtered, as it keeps going back and forth:

\begin{quote}
\itshape
 First, the scenario: There's a shoe box with kids' toys inside, but you can't see inside. The label says "nike shoes." Jake's friend told him that there are kids' toys inside and to ignore the label. Jake believed his friend. Then Jake finds the box, reads the label, and gets delighted because he's a shoe enthusiast.

The question is whether it's true or false that Jake is delighted.

Jake believed his friend, so he knows that there are kids' toys inside, not shoes. But when he sees the label, which says "nike shoes," he might forget or disregard what his friend said because he's a shoe enthusiast. However, the problem says that Jake believed his friend, so he should still believe that there are toys inside.

But the key point is that Jake is delighted when he reads the label. Since he believes his friend, he knows that the label is misleading, so he shouldn't be delighted by the label because he knows it's not true. But he is a shoe enthusiast, so the thought of shoes might excite him even if he knows it's not real.

(...)

However, the question is about his emotional reaction. It says he is delighted to find it. But if he knows it's not shoes, why would he be delighted? Unless he is delighted by the box itself or the possibility, but that seems unlikely.

(...)

So, based on belief, he knows there are toys inside, so he shouldn't be delighted about shoes.

Therefore, it might be false that he is delighted.

(...)

But the narrative explicitly says he is delighted, so from the story, it is true that he is delighted.

(...)

So, since he reads the label and is delighted, it is consistent that he is delighted because he is not ignoring the label.

Therefore, it is true that he is delighted.

(...)

Yes, that makes sense.

Therefore, Jake is delighted to find it, so True.

\end{quote}

\paragraph{Model gets it wrong.} The only cases where models get results completely wrong are in the modifications 2A and 2B. Here is an example of the reasoning summary of Claude in problem 2B.1:

\begin{quote}
\itshape
    Let me trace through this step by step:
    \begin{enumerate}
        \item Initial state: Adu, Lin, flower, suitcase, and chest are in the room
        \item Adu takes the flower and puts it on the chest
        \item Adu leaves for school
        \item While Adu is away, Lin moves the flower from the chest to the suitcase
        \item Lin leaves for swimming class
        \item Adu returns from school
    \end{enumerate}
    
    The key point here is that Adu doesn't know what happened while he was away. From Adu's perspective, the last thing he knew was that he put the flower on the chest. He has no knowledge that Lin moved it.
    
    This is a classic false belief task - Adu will have a false belief about where the flower is because he doesn't know it was moved. He will look for it where he left it.
    
    Therefore, Adu will look for the flower on the chest.
\end{quote}

One could see this and think that the model absorbed spurious correlations from the training data, which makes it respond to it incorrectly, as it even says that this is a 'classic false belief test'. We argue, however, that the difference from the "classic false belief test" in the two cases where models can't solve the tests (including this one) requires a mental visualization of the task for the difference to be trivial.
\subsection{Benchmark results}\label{benchResults}

\begin{table*}[!htbp]
\footnotesize
\setlength{\extrarowheight}{3.5pt}
\setlength{\tabcolsep}{3pt}
\begin{center}
\begin{tabular}{c c|c|c|c}
Model & ParaphrasedToMi & BigToM & FANToM & MMToM-QA \\ \hline
\begin{tabular}{c}
\\
   \\
   GPT-4o\\ \hline
   o1-preview\\
   o1-low-effort\\
   o1-medium-effort\\
   01-high-effort\\ \hline
   o1-mini \\
   o3-mini-low-effort \\
   o3-mini-medium-effort\\
   o3-mini-high-effort \\ \hline
   Deepseek R1 \\ \hline
   Llama 3.3 \\
   Gemini 1.5 Pro \\
   Qwen 2.5 72B \\
   QwQ 32B preview \\
   
\end{tabular} &
\begin{tabular}{p{7mm} p{9mm} p{9mm}}
   Avg. & False Belief & True Belief \\ \hline
   59.5 & 38.5 & 80.5 \\ \hline
   68.0 & 100.0 & 36 \\
   67.8 & 98.5 & 37.0 \\
   70.8 & 96.0 & 45.5 \\
   73.0 & 97.0 & 49.0\\ \hline
   60.0 & 60.0 & 60.0\\
   64.0 & 62.5 & 65.5\\
   64.5 & 90.5 & 38.5\\
   64.0 & 99.5 & 28.5\\ \hline
   68.3 & 77.0 & 59.5\\ \hline
   57.3 & 44.5 & 70.0\\
   54.8 & 43.5 & 66.0\\
   57.3 & 39.0 & 75.5\\
   58.8 & 27.5 & 90\\
\end{tabular} &
\begin{tabular}{c}
Avg.\\
\\ \hline
   98.0\\ \hline
   98.3\\
   99.2\\
    98.8\\
   98.8\\ \hline
   97.6 \\
   95.2\\
   97.2\\
   97.2\\ \hline
   96.8\\ \hline
   96.4\\
   98.0\\
   94.0\\
   93.6\\
   
\end{tabular} &
\begin{tabular}{p{9mm} p{11mm} p{13mm}}
   All Qs. & Ans. All Qs & Info Acc. All Qs \\ \hline
   11.1& 33.3& 32.7\\ \hline
   44.4& 66.7& 63.5\\
   28.3& 41.5& 51.0\\
   30.2& 41.5& 58.8\\
   37.9& 44.8& 63.0\\ \hline
   9.4& 41.5& 23.5\\
   0.0& 17.0& 11.8\\
   1.9& 26.6& 19.6\\
   1.9& 35.8& 33.3\\ \hline
   37.9& 62.1& 48.1\\ \hline
   0.0& 7.4& 0.0\\
   1.9& 7.5& 2.0\\
   0.0& 5.7& 2.0\\
   0.0& 0.0& 0.0\\
\end{tabular}&
\begin{tabular}{p{7mm} p{9mm} p{9mm}}
   All & Belief & Goal \\
    & & \\ \hline
   56.5& 74.5& 39.8\\ \hline
   70.4& 96.0& 51.0\\
   76.5& 96.1& 57.1\\
   76.0& 94.1& 60.2\\
   76.5& 95.1& 59.2\\ \hline
   60.0& 91.2& 27.6\\
   55.0& 71.6& 37.8\\
   69.0& 94.2& 42.9\\
   71.5& 97.1& 44.9\\ \hline
   49.0& 73.5& 24.5\\ \hline
   47.0& 50.0& 42.9\\
   50.0& 70.6& 28.6\\
   41.5& 51.0& 32.7\\
   51.5& 71.7& 29.6\\
\end{tabular}
\end{tabular}
\end{center}
\caption{Benchmark results of several models. Data from \cite{kimHypothesisDrivenTheoryofMindReasoning2025}}\label{benchmarks}
\end{table*}

It is possible to see in the table \ref{benchmarks} that the reasoning models perform significantly better than most non-reasoning ones for most benchmarks, with the exception of GPT-4o, which possesses comparable results at times. The benchmark in which the discrepancy is most notable is the FANToM Benchmark. It is hypothesized that this is due to the fact that for a model to score in this particular benchmark, it needs to correctly respond to \textbf{all} the questions of a particular type. This makes it so that the lack of consistency of non-reasoning models is accentuated in the results.

Additionally, it is possible to notice that the models of the distilled type have a significantly worse performance in this particular benchmark than the rest of the reasoning models. No satisfactory interpretation can be provided for that, as the exact characteristics of distilled models are still being discussed in the literature \cite{yueDoesReinforcementLearning2025}. Also, the authors of the original paper point to a correlation between effort and degradation in true-belief test performance, but this seems to be false for the o1 model, the only non-destilled model in the benchmark with variable effort. 

\section{Discussion}
%\label{Discussion, Limitations, and Further Research}
\label{Discussion}
In this section, we reflect on the meaning of the results given the literature (\ref{reflection}) and present the limitations of the present paper, suggesting ways to circumvent them with future research (\ref{limitations}).

\subsection{Reflection on Results}\label{reflection}

The results from \ref{PsiResults} demonstrate that models significantly improved their performance in ToM tasks since 2023. Here, we argue that the gains displayed by the models are mainly due to an overall increase in how powerful models are, but partly due to a newly found robustness of their ToM skills, which may be explained partly due to their reasoning abilities. \cite{yueDoesReinforcementLearning2025} demonstrates that the reasoning paths achieved by reasoning models are bound by the original capabilities of their base models, achieving asymptotic performance in tasks without increasing the perplexity on the base model. Here it is argued that, in practice,  RLVR-trained models are more robust to differences in prompting. Across experiments, the primary effect of RLVR training appears to be improved stability in inference paths rather than expanded representational capacity

The obtained results in the psychological tests indicate that new models can be incredibly robust in tasks. For instance, in all psychological tests taken from \cite{duijnTheoryMindLarge2023}, GPT-5 made only one partial mistake. It was also shown that performance in the original 'modifications' experiment was significantly increased in the Claude model when thinking is on comparing with its non-thinking counterpart. This increase in performance is attributed to a gain in robustness due to its ability to use RLVR-trained inference time scaling. It is speculated that, in a test where prompts are specifically designed to disturb the reasoning of models, this feature makes more of a difference. 

Two of the 'modification' tasks, however, weren't successfully completed by a single model. This is possibly because those modifications are not as trivial as one might have anticipated, as they might require picturing the scene: for humans, picturing scenes in our minds is so common that it is natural to forget how essential of a skill it is in certain contexts.

It is argued that the trivial modifications 2A: Transparent Access and  2B: Relationship Change (see \ref{Appendix A}) are not really trivial, as they require a visual mental depiction of the scene for the understanding that the objects being inside a transparent recipient or on something (rather than in something) make the information of where they are immediately available to the viewer when they look in the direction of the object. Of course, one can still, through many steps, derive this semantically, but it becomes a substantially harder task than anticipated.

It is also interesting to notice that one of the only two stories in the 'Strange Stories' psychological test in which models made a partial mistake had a visual aspect to it that could clarify which of the two hypothesis models considered in their reasoning is the correct one. 

As seen in subsection \ref{ToM and LLMs}, through the years, different prompting strategies have successfully emerged to induce LLMs to perform better in ToM tasks. It is hereby argued that this indicates that, although models had the knowledge and capabilities required to perform well in such tasks, they had to be induced to find the correct reasoning and response paths.

It is hypothesized that RLVR-trained reasoning models converged to use some locally good inference-time scaling techniques, which improved their robustness in ToM tasks. This is also corroborated by the displayed results, which demonstrate that models even sometimes display a version of what the perspective-taking prompting technique tries to achieve without being provoked to do so.

This is also in line with the findings of \cite{ullmanLargeLanguageModels2023}, which showed that simple prompt alterations would lead the models to fail in the same types of tasks. Here we argue that this means that the upper bound for success was already present in the models, not that they didn't possess 'real' ToM skills. In other words, they already, to a large extent, had the necessary skills to solve ToM problems, but weird prompts could easily disturb their reasoning.

To corroborate that, we analyzed the benchmark results from \cite{kimHypothesisDrivenTheoryofMindReasoning2025}, which, although featuring slightly older reasoning models, demonstrate that, in general, they scored significantly above their non-reasoning peers, suggesting a positive impact from reasoning in the solving of ToM tasks. Additionally, in the benchmark for which a lack of robustness was punished the most, even more pronounced differences were verified.  

With this in mind, the result from \cite{yueDoesReinforcementLearning2025} is hereby reinterpreted, at least in the context of ToM tasks. By understanding what reasoning models don't do, it is also possible to understand what they do. More generally, we argue that the narrowing of the reasoning coverage of RLVR models might be precisely their strength. While one might want models with very flexible results for some applications, it's generally desirable for models to be robust.

\subsection{Limitations and further research}\label{limitations}

Although evidence from our experiments and benchmarks in literature were found of reasoning helping models in answering queries more robustly in the context of ToM tasks, no clear quantification of such an effect can be provided with only this work. Future research could tackle such a problem by running ToM benchmark tests comparing RLVR-trained models with their base model counterparts using the pass@k metric proposed by \cite{yueDoesReinforcementLearning2025}.  

Additionally, spurious correlations like relating the semantic structure of the tests with meta-knowledge of ToM can't be ruled out for all psychological test cases, which might make some of the results inconclusive for real-world scenarios.

The present work also does not directly compare the RLVR-trained models' performance in ToM with base models employing other inference-time scaling techniques; the extent to which the increase in performance can be attributed directly to this specific way of increasing inference time can be investigated in future work.

Furthermore, some of the evidence motivating the conclusions of this paper is drawn by comparing the performance of Claude with thinking on with its counterpart with thinking off. Claude, with thinking off is not the same thing as its base model, which is not accessible, yet we hope the difference in performance sheds some light on what the reasoning process is actually helping the model with. Further research could investigate differences in performance between reasoning models and base models in cases where they are available, such as R1 and V3-base.

It's also noteworthy that, although the reasoning traces were analyzed for strategies employed by the models, it has been argued that reasoning traces don't always present the real motivation behind a model's final response \cite{chenReasoningModelsDont2025}. 

Lastly, the different treatment between models might induce differences in performance. For example, setting the models' temperature to 0 or asking why a response was given directly affects the performance of the models in the given tests.

\section{Conclusions}\label{conclusions}

The present work discussed how RLVR-trained reasoning models perform in ToM tasks. Through the analysis of psychological experiments, including an original one, and substantiated by benchmark results from the literature, it was hypothesized that the advantage that such models have compared to their base models is an increase in robustness, rather than new reasoning paths. Such a result is compatible with literature findings and might re-signify some earlier findings, which argue that a lack of robustness in ToM tasks shows that LLMs have no 'real' ToM skills.

More generally, the paper also shows that models in general perform much better in those tasks than they did two years ago and that an increase in robustness when compared to the base model might be the real power of RLVR training.
\bibliographystyle{ACM-Reference-Format}
\bibliography{bibfile}

\appendix

\section{'Trivial' Modifications}\label{Appendix A}
In this appendix section, the new variations on the tests used in \cite{ullmanLargeLanguageModels2023} are presented.
\subsection{Variation 1A: Transparent Access}

There is a shoe box, inside of it there are kids toys. The box is made out of see-through cardboard, so you can see what is inside. Yet, there is "nike shoes" printed on the label. Jake finds the shoe box, which he had never seen before. He reads the label. 
Jake is a shoe enthusiast, he is delighted to find it. True/False?
He believes that the box contains ...

\subsection{Variation 1B: Uninformative Label}

Jake is on the road, he wants to find his way to Leiden but doesn't know the directions. He sees a sign that points out that Amsterdam is to the right and Leiden is to the left. Jake does not know how to read.
\begin{enumerate}
    \item Jake is now relieved. True/False?
    \item To which direction Jake will turn his car?
\end{enumerate}

\subsection{Variation 1C: Trustworthy Testimony}

There is a shoe box, inside of it there are kids toys but you can't see what is inside.  Yet, there is "nike shoes" printed on the label. Before finding the box, Jake's friend told him "there are kids toys inside, ignore the label". Jake believed his friend. Jake finds the shoe box, which he had never seen before. He reads the label, which says 'nike shoes' in it. 
\begin{enumerate}
    \item Jake is a shoe enthusiast, he is delighted to find it. True/False?
    \item He believes that the box contains ...
\end{enumerate}
\subsection{Variation 1D: The Treachery of Late Labels}

There is a shoe box, Jake puts kids toys inside of it. After that he closes the box and writes a label in it. He can't see what is inside the bag. He then looks at the box and reads 'nike shoes' . 
\begin{enumerate}
    \item Jake is a shoe enthusiast, he is delighted to find it. True/False?
    \item He believes that the box contains ...
\end{enumerate}

\subsection{Variation 2A: Transparent Access}

In the room there are Adu, Lin, a flower, an aquarium, and a transparent plastic bag with water. Adu takes the flower and puts it in the bag. He leaves the room and goes to school. While Adu is away, Lin takes the flower out of the bag and puts it in the aquarium. Lin leaves the room and goes to his swimming class. Adu comes back from school and enters the room. He doesn’t know what happened in the room when he was away.
 \begin{enumerate}
     \item Adu will look for the flower in the ...
     \item Adu thinks the flower is in the ...
 \end{enumerate}

\subsection{Variation 2B: Relationship Change}

In the room there are Adu, Lin, a flower, a suitcase, and a chest. Adu takes the flower and puts it on the chest. He leaves the room and goes to school. While Adu is away, Lin takes the flower off the chest and puts it on the suitcase. Lin leaves the room and goes to his swimming class. Adu comes back from school and enters the room. He doesn’t know what happened in the room when he was away.
 \begin{enumerate}
     \item Adu will look for the flower on the ...
     \item Adu thinks the flower is on the ...
 \end{enumerate}

\subsection{Variation 2C: Trusted Communication}

In the room there are Adu, Lin, a flower, a suitcase, and a chest. Adu takes the flower and puts it in the chest. He leaves the room and goes to school. While Adu is away, Lin calls Adu to tell him he is going to move the flower to the suitcase. Adu believes him. While Adu is away, Lin takes the flower out of the chest and puts it in the suitcase. Lin leaves the room and goes to work. Adu comes back from school and enters the room. He doesn’t know what happened in the room when he was away.

\subsection{Variation 2D: Querying the Mental States of the Additional Person}

In the room there are Adu, Lin, a flower, a suitcase, and a chest. Adu takes the flower and puts it in the chest. He leaves the room and goes to school. While Adu is away, Lin takes the flower out of the chest and puts it in the suitcase. Lin leaves the room and goes to his swimming class. Lin and Adu come back and enter the room. They don’t know what happened in the room when they were away.

 \begin{enumerate}
     \item Lin will look for the flower in the ...
     \item Lin thinks the flower is in the ...
 \end{enumerate}

\end{document}